\documentclass[10pt,a4paper]{article}
\usepackage[margin=1in]{geometry}
\usepackage[numbers]{natbib}
\usepackage{amssymb}
\usepackage{amsfonts}
\usepackage{mathrsfs}
\usepackage{amsmath}
\usepackage{babel}
\usepackage{lastpage}
\usepackage[utf8]{inputenc}
\usepackage[bookmarks=true,colorlinks=true,linkcolor=blue,citecolor=blue]{hyperref}
\usepackage{graphicx}
\usepackage{fancyhdr,enumerate}
\usepackage{color}
\usepackage[mathlines]{lineno}
\usepackage{lscape}
\usepackage{natbib}
\usepackage{subfigure}
\usepackage{float,adjustbox}
\usepackage{bm}
\usepackage{booktabs}
\usepackage[margin=1in]{geometry}

\usepackage{url} 
\usepackage{booktabs}
\usepackage{setspace}
\usepackage{lmodern} 
\usepackage{array} 

\newcolumntype{P}[1]{>{\raggedright\arraybackslash}p{#1}}

\begin{document}

\title{Classification and Prediction of Heart Diseases using Machine Learning Algorithms}

\author{Akua Sekyiwaa Osei-Nkwantabisa $^{1,2}$ and Redeemer Ntumy $^{3}$\\
 $^{\textbf{1}}$	School of Mathematical and Statistical Sciences,\\ College of Sciences, University of Texas Rio Grande Valley, USA.\\
 $^\textbf{{2}}$Department of Statistics and Actuarial Science,\\ College of Basic and Applied Sciences, University of Ghana, Ghana.  \\
 $^\textbf{{3}}$ Department of Computer Science,\\College of Basic and Applied Sciences, University of Ghana, Ghana.  \\}

\date{}
\maketitle
\begin{abstract}\medskip
	\noindent
	Heart disease is a serious worldwide health issue because it claims the lives of many people who might have been treated if the disease had been identified earlier. The leading cause of death in the world is cardiovascular disease, usually referred to as heart disease. Creating reliable, effective, and precise predictions for these diseases is one of the biggest issues facing the medical world today. Although there are tools for predicting heart diseases, they are either expensive or challenging to apply for determining a patient's risk. The best classifier for foretelling and spotting heart disease was the aim of this research. This experiment examined a range of machine learning approaches, including Logistic Regression, K-Nearest Neighbor, Support Vector Machine, and Artificial Neural Networks, to determine which machine learning algorithm was most effective at predicting heart diseases. One of the most often utilized data sets for this purpose, the UCI heart disease repository provided the data set for this study. The K-Nearest Neighbor technique was shown to be the most effective machine learning algorithm for determining whether a patient has heart disease. It will be beneficial to conduct further studies on the application of additional machine learning algorithms for heart disease prediction.\\
	
	
\end{abstract}

\newpage

\section{Introduction}
The heart and blood arteries are affected by a group of ailments known as cardiovascular diseases (CVDs) and cardiac disorders. These conditions include deep vein thrombosis, coronary heart disease, peripheral arterial disease, and rheumatic heart disease. One of our body's most important organs is the heart. It is a muscular organ found directly beneath and just to the left of the breastbone. The World Health Organization estimates that CVDs account for 17.9 million deaths annually, making them the leading cause of death worldwide. The majority of these fatalities, which make up 85\% of them and happen in low- and middle-income nations, are mostly caused by heart attacks and strokes. Furthermore, persons under the age of 70 account for one-third of these early mortality. Poor diet, inactivity, smoking, and binge drinking are a few risk factors for heart disease and stroke. These psychological risk factors can manifest physically as obesity, overweight, high blood lipid levels, high blood pressure, and high blood sugar. In order to take preventative actions and avoid the damage that these disorders can cause, early identification of cardiovascular diseases is essential. Underlying blood vessel issues may go untreated for a long time. The condition typically manifests first as heart attacks and strokes. Heart attack symptoms include pain or discomfort in the middle of the chest, in the arms, left shoulder, elbow, jaw, or back. The person may also have back or jaw discomfort, nausea or vomiting, lightheadedness, or fainting. Common stroke symptoms include trouble speaking or understanding speech, disorientation, difficulty seeing with one or both eyes, numbness on one side of the face, arm, or leg, difficulty walking, and a severe headache with no known reason. Due to a lack of basic healthcare facilities for early detection and treatment, cardiovascular diseases often have higher death rates in low and middle-income nations. People with CVDs have limited access to adequate, equitable healthcare services that can meet their needs in low and middle-income countries. Due to the disease's delayed identification, people get CVDs and pass away early. It has been demonstrated that lowering salt intake, increasing fruit and vegetable consumption, engaging in regular exercise, quitting smoking, and abstaining from excessive alcohol consumption all reduce the risk of cardiovascular disease. Additionally, recognizing people who are most at risk for CVDs and ensuring they receive the right care will help avoid early mortality. All primary healthcare facilities must have access to necessary medications and fundamental medical technology to guarantee that individuals in need receive treatment and counseling \citep{who2021cardiovascular}.\\

\noindent
Machine learning is becoming more and more popular as its value in different industries increases daily. Among the industries that apply machine learning include manufacturing, retail, healthcare, life sciences, travel and hospitality, financial services, energy, feedstock, and utilities. The healthcare sector is one of these applications' most significant industries \citep{akhila2022prediction}. The area of machine learning is expanding quickly as a result of the massive volumes of data being collected. Many big data experts predict that the volume of data generated will continue to rise sharply in the future. The International Data Corporation (IDC) projects that the world's datasphere will grow to 175 zettabytes by 2025 in its Data Age 2025 research. To put this into context, the stack would have covered two-thirds of the distance between the Earth and the Moon in 2013 if we converted this amount to 128GB iPads. This stack would be 26 times longer in 2025 \citep{khvoynitskaya2020future}. Given this, it is crucial to comprehend data and obtain insights to comprehend the world of humans. Compared to human doctors, machine learning is more accurate and quicker at diagnosing. In the face of uncertainty, machine learning algorithms create models that make predictions based on data. These techniques train a model to produce accurate predictions using known sets of inputs and outputs. The biggest issue facing the medical industry today is making accurate and dependable predictions for disease diagnosis and treatment. Although there are techniques for forecasting cardiovascular diseases, they are usually expensive or ineffective at determining an individual's risk. Early detection of CVDs can considerably reduce the chance of fatalities and other issues associated to these conditions. Using data mining and machine learning approaches, automation can help solve the issue of low prediction accuracy. Data mining searches through massive data sets using a range of computational technologies in order to identify patterns and predict outcomes. In order to address the issue of poor CVD prediction accuracy, the goal of this project is to design and put into practice a system for heart disease detection and prediction utilizing machine learning techniques. Finding the best classifier for predicting and diagnosing cardiovascular diseases is the major goal of this study. Based on factors including gender, age, cholesterol levels, and other medical traits, the goal is to ascertain whether a patient has a cardiovascular disease. The system looks for and extracts helpful insights from prior data sets to aid in the diagnosis and prevention of heart disease.

\newpage
\section{Literature Review}
\vspace{-0.4em}
Machine learning is a technique for automatically identifying patterns in data. Supervised, unsupervised, and reinforcement learning are the three main categories of machine learning. A technique called supervised learning, commonly referred to as supervised machine learning, uses labeled training data to learn how to predict results for new data.Support vector machines, decision trees, logistic regression, and naive bayes classifiers are a few supervised machine learning techniques. It is essential to identify and anticipate heart disorders early because doing so can lower mortality rates and overall consequences. As a result, various studies on the topic of heart disease have been carried out using data mining and machine learning approaches. In order to improve the precision of the prediction of cardiovascular disease, \citep{mohan2019effective} suggested a unique method that makes an effort to identify crucial variables by utilizing machine learning techniques. Using the UCI heart disease data set, the accuracy of the recommended strategy was compared to various machine learning techniques. A unique technique known as Hybrid Random Forest with Linear Model (HRFLM) was developed for this investigation. In their hybrid HRFLM technique, the authors combined the traits of Random Forest (RF) and Linear Method (LM). HRFLM surpassed Decision Tree (DT), Random Forest (RF), and Linear Method (LM) in terms of the number of attributes and prediction error, demonstrating that it is quite accurate in predicting heart disease. In order to improve the accuracy of heart disease prediction and gain a deeper understanding of the key features, new feature-selection strategies can be developed, claim \citep{mohan2019effective}. Several machine learning techniques were employed by \citep{saboor2022method} in their study to identify and forecast cardiovascular disease. The algorithms' performance was then evaluated using a variety of metrics, such as classification accuracy, sensitivity, specificity, and F measure, on the heart disease data set. Nine (9) machine learning classifiers were applied to the final data set both before and after the hyper parameter tuning. These included the AdaBoost Classifier (AB), the Logistic Regression (LR), the Extra Trees Classifier (ET), the Multinomial Naive Bayes (MNB), the Decision Trees (CART), the Support Vector Machine (SVM), the Linear Discriminant Analysis (LDA), the Random Forest (RF), and the XGBoost (XGB). Online repositories such as the Cleveland heart disease data set, Z-Alizadeh Sani data set, Statlog Heart, Hungarian Long Beach VA, and Kaggle Framingham data set were the sources of the data sets used by \citep{saboor2022method}. In conclusion, a variety of classifiers were employed to forecast the development of heart disease, with the Support Vector Machine emerging as the most reliable. In this study article by \citep{saboor2022method}, some disadvantages include the fact that the functioning of the earlier proposed systems is significantly lowered if the size of the data set is raised. Additionally, the classifier prediction accuracy increases as the size of the data set grows, but beyond a certain size, the accuracy of the classifier prediction decreases.\citep{haq2018hybrid} created a machine learning-based diagnosis method for heart disease prediction using a data set of heart disease. Cross-validation, three feature selection techniques, seven well-known machine learning algorithms, and metrics for classifier performance assessment such as classification accuracy, specificity, sensitivity, Mathews' correlation coefficient, and execution time were all used. This study employed the Cleveland heart disease data collection from 2016, which is popular among researchers. A hybrid intelligent machine learning-based prediction system was proposed for the diagnosis of heart disease. The seven well-known classifiers Logistic Regression, K Nearest Neighbor, Artificial Neural Network (ANN), Support Vector Machine (SVM), Naive Bayes, Decision Tree, and Random Forest were used to select the critical features using the three feature selection algorithms Relief, Minimal-Redundancy-Maximum-Relevance Feature Selection Algorithm (MRMR), and Least Absolute Shrinkage and Selection Operator (LASSO). Logistic regression with 10-fold cross-validation demonstrated the best accuracy when it was selected by the FS algorithm Relief. However, in terms of specificity, the MRMR algorithm surpassed SVM (linear) with feature selection. The authors concluded that more research is required \citep{haq2018hybrid}, to improve the efficiency of these predictive classifiers for the diagnosis of heart disease by using various feature selection algorithms and optimization strategies.In their work \citep{vardhan2022heart} established a technique for detecting the existence of heart disease using clinical data collected from subjects. The main objective was to create a predictive model for heart disease using a variety of characteristics. Different machine learning classification strategies were also tested and assessed using traditional performance metrics like accuracy in order to compare various machine learning algorithms. For this experiment, data from the UCI Heart Disease Data Collection were used. Machine learning methods such Random Forest, Support Vector Machine, K-Nearest Neighbor, Decision Tree, Artificial Neural Networks, Logistic Regression, and Naive Bayes were used to compare how well they predicted heart diseases.The Random Forest algorithm was shown to be the most reliable strategy for predicting cardiac illness, with an accuracy rate of 90.16 percent, according to \citep{vardhan2022heart}. In their conclusion, \citep{vardhan2022heart} recommended using large data sets in trials in the future to increase the reliability of their findings and help doctors better anticipate heart disease.From these related works, it shows that machine learning plays a critical role in the classification and prediction of heart disease.

\newpage
 \section{Data and Methods}
 \noindent
 This suggested model's data came from the UCI Machine Learning Repository as its data source. The Heart Disease Data Set was employed, and it has been a popular global resource for students, instructors, and researchers looking for machine learning data sets. David Aha and graduate students at UC Irvine produced the data set in 1987. Four datasets from different institutions make up the heart disease data set, including  Cleveland ,Hungarian ,Switzerland, Long Beach VA and Statlog (Heart) Data Set.The Heart Disease Data Set has 76 attributes and is made up of four databases from different universities. Only 12 of these traits—including the projected trait are utilized in this investigation.The target field in the data set indicates whether a patient has heart disease or not, with 0 signifying no disease and 1 signifying the presence of a disease. The form of the data collection is (1190,12).
 
 \begin{table}[!htbp]
    \centering
    {\small
     \caption{Feature Description}
    \begin{tabular}{| P{0.3in} | P{1.2in} | P{3.0in} | P{0.7in} |} \hline
       \textbf{No.} & \textbf{Feature} & \textbf{Description} & \textbf{Data type} \\ \hline
        1. & Age & Patients' years of age & Numeric \\ \hline
       2. & Sex & Gender of Patient (Male - 1, Female - 0) & Nominal \\ \hline
      3. & Chest pain type & Chest pain type & Nominal \\
      & & 1 = Typical angina & \\
      & &  2 = Atypical angina & \\
      & &  3 = Non-anginal pain & \\
      & &  4 = Asymptomatic  & \\ \hline
4. & Resting BP & Level of blood pressure at resting mode in mm/HG & Numeric\\ \hline 
5. & Cholesterol & Serum cholesterol in mg/dl & Numeric\\ \hline
6. & Fasting blood sugar & Fasting blood sugar & Nominal \\
     & & 0 = Less than 120 mg/dl & \\
    & & 1 = More than 120 mg/dl & \\ \hline 
7. & Resting ECG & Resting electrographic result &  Nominal \\
  & & 0 = Normal & \\
    & & 1 = Having ST-T wave abnormality & \\
    & & 2 = left ventricular hypertrophy & \\ \hline
8. & Max Heart rate &  Maximum heart rate achieved & Numeric \\ \hline
9. & Exercise angina & Exercise induced angina & Nominal \\
    & & 0 = No & \\
    & & 1 = Yes & \\  \hline 
10. & Old peak & Exercise induced ST-depression in comparison with the state of rest & Numeric\\ \hline

11. & ST slope & Slope of the peak exercise ST segment & Nominal\\
& & 0 = Normal & \\
& & 1 = Unsloping & \\
& & 2 = Flat & \\
& & 3 = Downsloping & \\ \hline
 \multicolumn{4}{|c|}{\textbf{TARGET VARIABLE}} \\
 \hline
12. & Target & It is the target variable that we must forecast; a value of one denotes a patient who is at heart risk, whereas a value of zero indicates a healthy patient  & Heart Risk, Nominal\\
\bottomrule
    \end{tabular}
    }
    \label{tab:feature_description}
\end{table}
\vspace{0.2in}
\noindent
The machine learning algorithms used were K-Nearest Neighbors, Support Vector Machine, Logistic Regression and Artificial Neural Networks((Tensorflow). Before analysis was done on the data, the dataset was broken to training ($80\%$, $n =952$) and testing ($20\%$, $n = 238$).The training dataset was used to build and train the machine learning model while the testing data set was used  to evaluate its performance. Evaluating the performance of a machine learning model is one of the crucial steps in its development. To evaluate the efficacy  of the model, evaluation measures, sometimes referred to as performance metrics, are employed. These metrics enable us to assess how well the model processed the supplied data. The five metrics used were Confusion Matrix, Precision, Recall, F1- Score and Classification Accuracy.Hyper-parameter tuning and optimization was then performed to the models to help increase accuracy and determine the best model for predicting heart diseases. GridSearchCV,  a frequent method for adjusting hyperparameters was used in this project. For each combination of the provided hyper parameter values, a model is created, evaluated, and the design that yields the best results is chosen \citep{jordan2017hyperparameter}.

\newpage
\section{Results and Discussion}
\vspace{-0.5em}
The results of data analysis utilizing several machine learning models, such as K-Nearest Neighbors, Support Vector Machine, Logistic Regression, and Artificial Neural Network (Tensorflow), are presented in this chapter. Using the UCI data to forecast heart disease, these models' performance was assessed. Hyper parameter tuning was also done to improve the models' ability to accurately predict heart disease in patients under particular circumstances. 80 percent of the data was used for training, while 20 percent was used for testing. The data contained no missing values, and Python was the study's chosen programming language.
\vspace{-0.5em}
\subsection{Graphical Summary of UCI Data set}
\subsection*{Qualitative Variables}
\vspace{-0.5em}
\begin{figure}[!htbp]
      \centering
      \includegraphics[height=2.5in,width=6in]{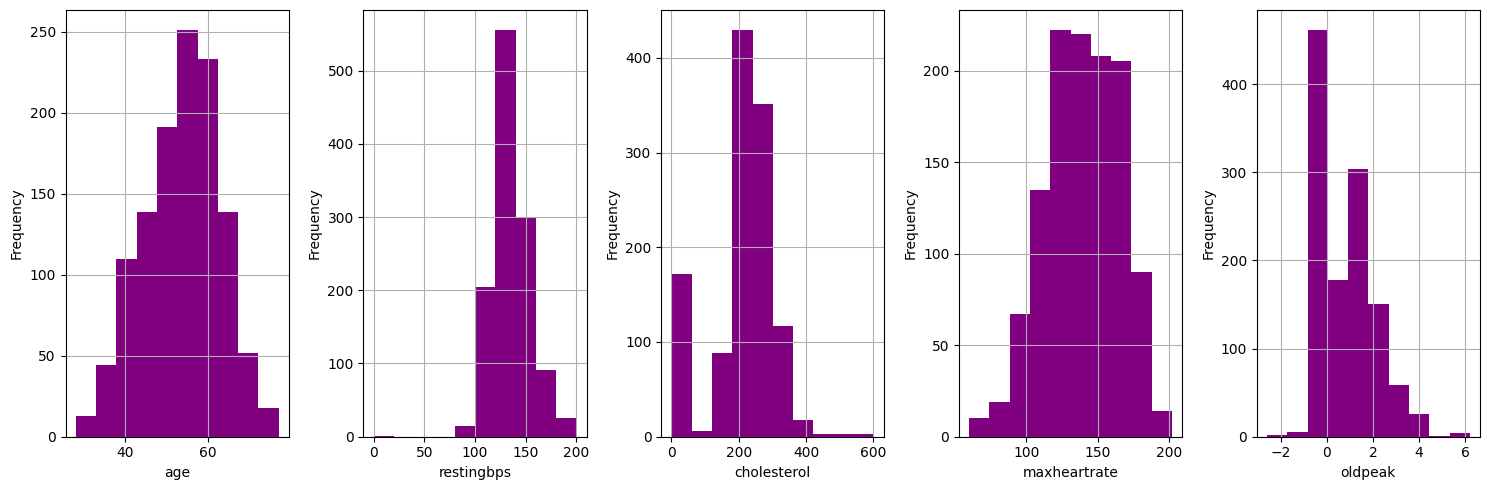}
      \caption{Subplots of the Qualitative Varibles}
    \end{figure}
\vspace{-0.5em}
\noindent
From the figure above, age and maximum heart rate are normally distributed hence can be used directly. Resting blood pressure and cholesterol may require transformations for improved modeling results because they are both skewed to the right. The skewness of oldpeak indicates that it must be handled delicately so as not to introduce any type of bias.

\subsection*{Quantitative Variables}
\vspace{-0.5em}
\begin{figure}[!htbp]
      \centering
      \includegraphics[height=2.5in,width=6in]{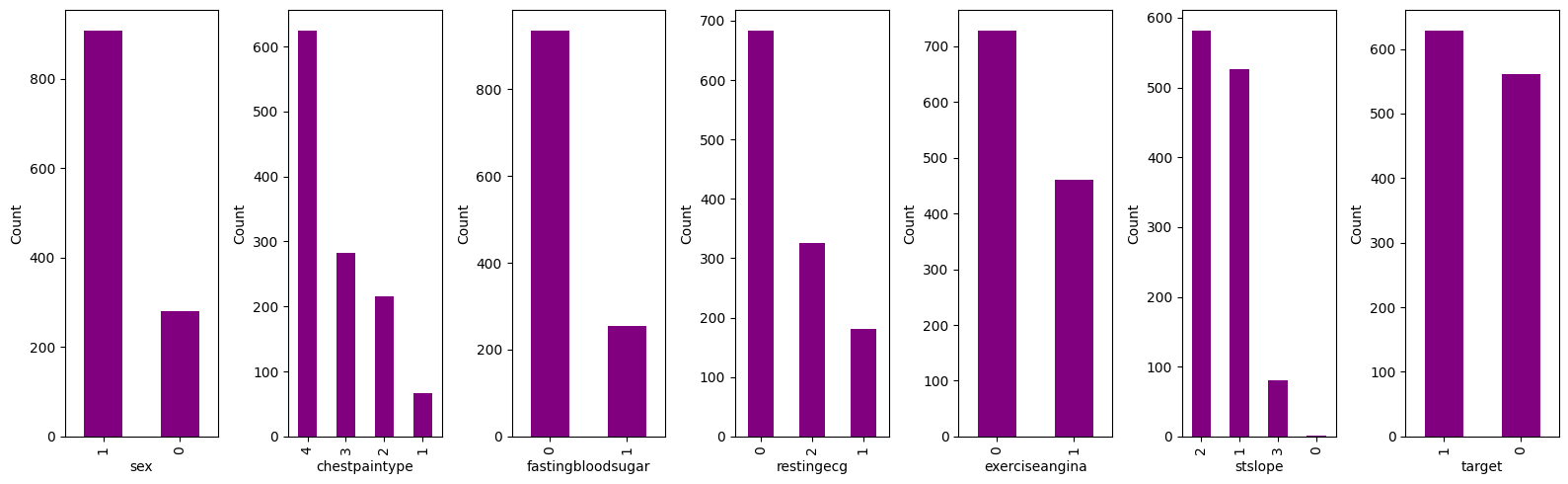}
      \caption{Subplots of the Quantitative Variables}
    \end{figure}

\noindent
It is evident that the imbalances here can affect the model by potentially biasing it towards the majority classes,which may necessitate the use of resampling or weighting methods for ensuring fair learning and accurate predictions across all categories.
    
\newpage
\subsection*{Correlation Matrix}
\begin{figure}[!htbp]
      \centering
      \includegraphics[height=3in,width=5in]{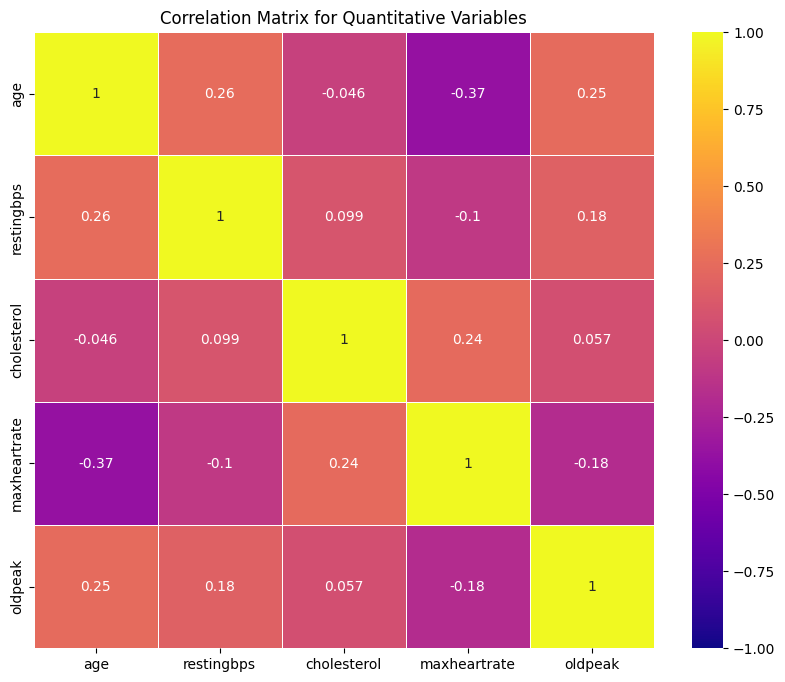}
      \caption{Correlation Matrix for Quantitative Variables}
    \end{figure}

\noindent
The correlations imply that in predicting heart disease, age, resting blood pressure and maximum heart rate may give more insight to the model to have a higher predictive power as compared to cholesterol. These stronger correlations should guide the modeling process while also considering other methodologies like feature engineering as well as interaction variables which aim at capturing these relationships properly.

\subsection{Heart Disease Frequency for Sex}
\vspace{-0.5em}
\begin{figure}[!htbp]
      \centering
      \includegraphics[height=3.5in,width=5in]{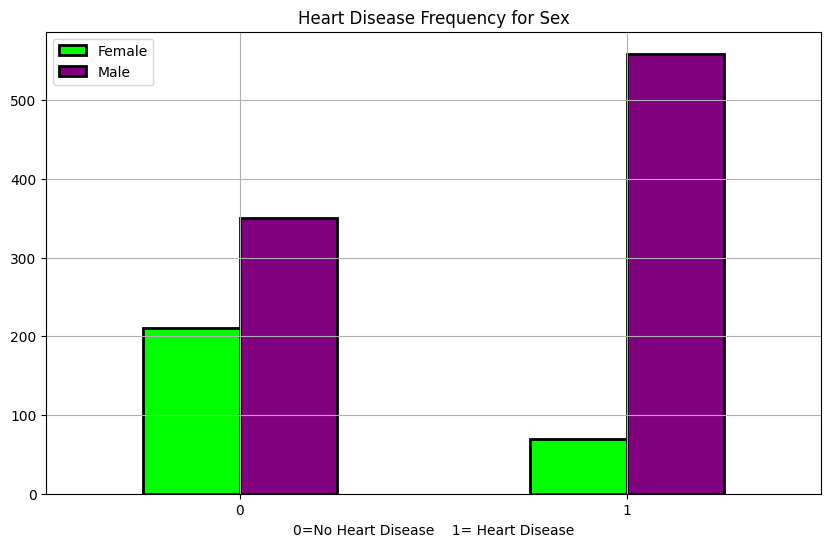}
      \caption{Heart Disease Frequency for Sex}
    \end{figure}
    
\noindent
As seen in Figure 4, the frequency of heart disease is displayed by sex. Per the chart, it is evident that the data is highly dominated by the male sex. This imbalance is likely to negatively impact the model's performance.
\newpage
\subsection{Performance of Machine Learning Algorithms}

\begin{figure}[!htbp]
  \centering
  \begin{minipage}{0.49\textwidth}
    \centering 
    
    \includegraphics[width=\textwidth]{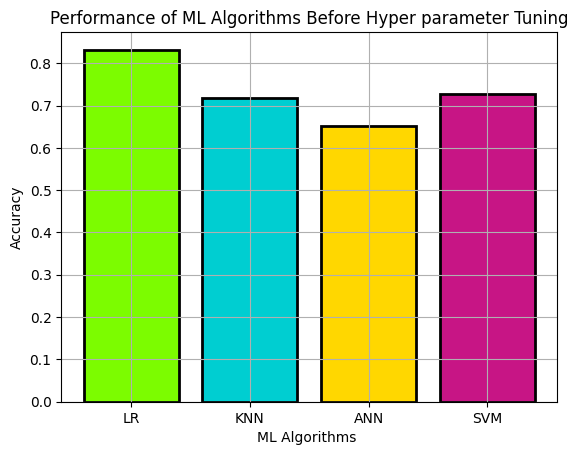}
    \caption{\centering Performance of Machine Learning Algorithms Before Hyper-Parameter Tuning}
  \end{minipage}
  \hfill
  \begin{minipage}{0.49\textwidth}
    \centering
    
    \includegraphics[width=\textwidth]{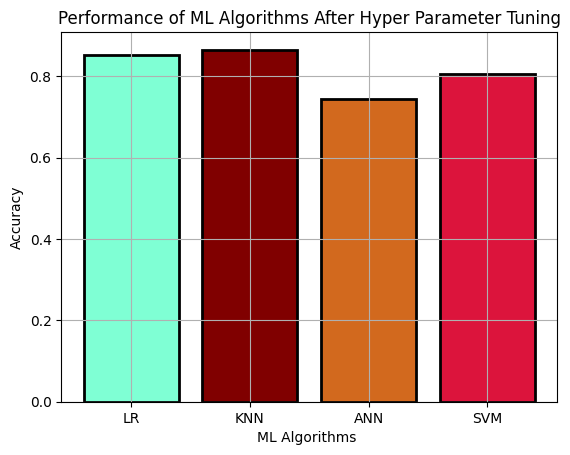}
   \caption{ \centering Performance of Machine Learning Algorithms After Hyper-Parameter Tuning}
  \end{minipage}
  \vspace{1cm} 
\end{figure}
\vspace{-0.5em}

\noindent
Before tuning the model, Figure 5 shows the output of our baseline models. The model with the highest accuracy, almost 83\%, was logistic regression. The K-Nearest Neighbors and Support Vector Machine algorithms came next, with accuracy rates of 72\% and 73\%, respectively. The accuracy of the tensorflow model was 65\%.\\

\noindent
The performance was significantly enhanced after model tuning. The Logistic Regression, Support Vector Machine, and K-Nearest Neighbors algorithms were tuned using GridSearchCV.\\

\noindent
The tensorflow model on the other hand, had some strategic adjustments made. These included L2 regularization and leakyReLU activations which would prevent overfitting and enhance learning dynamics. Dropout rates are tuned from 0.5 to 0.2 and Adam optimizer learning rate was fine-tuned at 0.0005. More patience was added for early stopping (20 epochs) and a ReduceLROnPlateau callback was inserted for adjusting learning rate based on validation loss. The batch size was adjusted to 20, improving gradient estimates as well as convergence speed.

\noindent
Figure 6 shows that, with the use of hyper parameter adjustment, all algorithms have an accuracy of over 80\% except Artificial Neural Network. The Support Vector Machine (SVM) performs well for predicting cardiac disease in patients despite having the lowest accuracy, which is 81\%. While Logistic Regression has an accuracy of 86 percent, Artificial Neural Networks (ANN) have an accuracy of 74 percent. The K-Nearest Neighbors algorithm has an 87 percent accuracy.

\subsection{Performance Evaluations of Machine Learning Algorithms}

    The models were evaluated according to these four (4) measures; \textbf{Accuracy}, \textbf{Precision}, \textbf{Recall} and \textbf{F1-score}.\\ 

\begin{table}[!htpb]
    \centering
   \vspace{-0.5em}
     \caption{Performance Metrics of the Machine Learning Algorithms}
    \begin{tabular}{P{1.6in} P{0.7in} P{0.7in} P{0.6in} P{1.0in}} \toprule
       Machine Learning Algorithms & Accuracy & Precision & Recall & F1-Score\\ \hline
       Logistic Regression & 0.85 & 0.87 & 0.86 & 0.87  \\
       Support Vector Machine & 0.81 & 0.81 & 0.84 & 0.83  \\
       K - Nearest Neighbors & 0.87 & 0.86 & 0.90 & 0.88  \\ 
       Artificial Neural Networks & 0.74 & 0.78 & 0.74 & 0.76  \\
       \bottomrule
    \end{tabular} 
   
    \label{tab:my_label}
\end{table}
\newpage

\begin{figure}[!htbp]
      \centering
      \includegraphics[height=3in,width=5in]{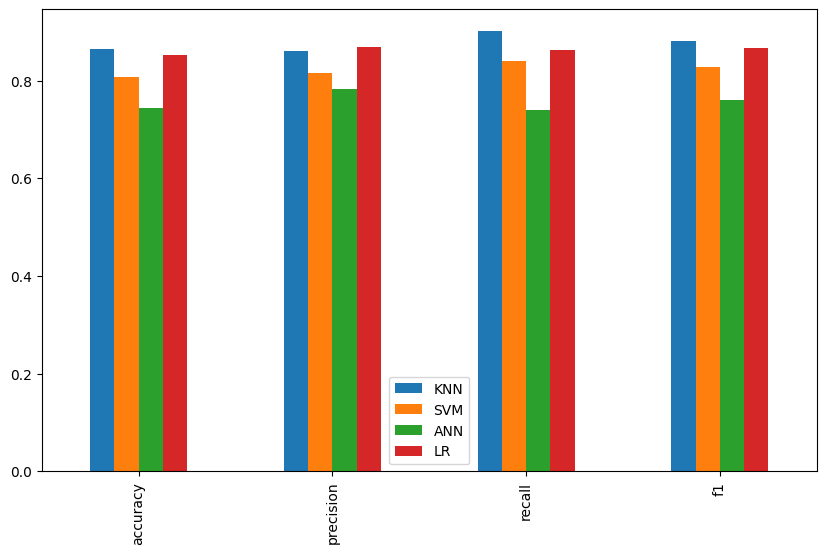}
      \caption{A Graph of the Performance Metrics of the Machine Learning Algorithms}
    \end{figure}

\noindent
 Table 4.2 and Figure 7 both reveal that the K-Nearest Neighbor algorithm delivered the greatest results, with accuracy rates of 87\%, precision rates of 86\%, recall rates of 90\%, and F1-Score rates of 88\%.

\noindent
The least performing model, tensorflow had an accuracy of 74 percent, precision of 78 percent, recall of 74 percent, and F1-Score of 76 percent.
\vspace{-0.5em}
\subsection{Confusion Matrix}
\noindent
A useful tool for providing more insight in how best a model predicts certain classes is the confusion matrix. The confusion matrices for the models are shown below

\begin{figure}[!htbp]
      \centering
      \includegraphics[height=4in,width=4in]{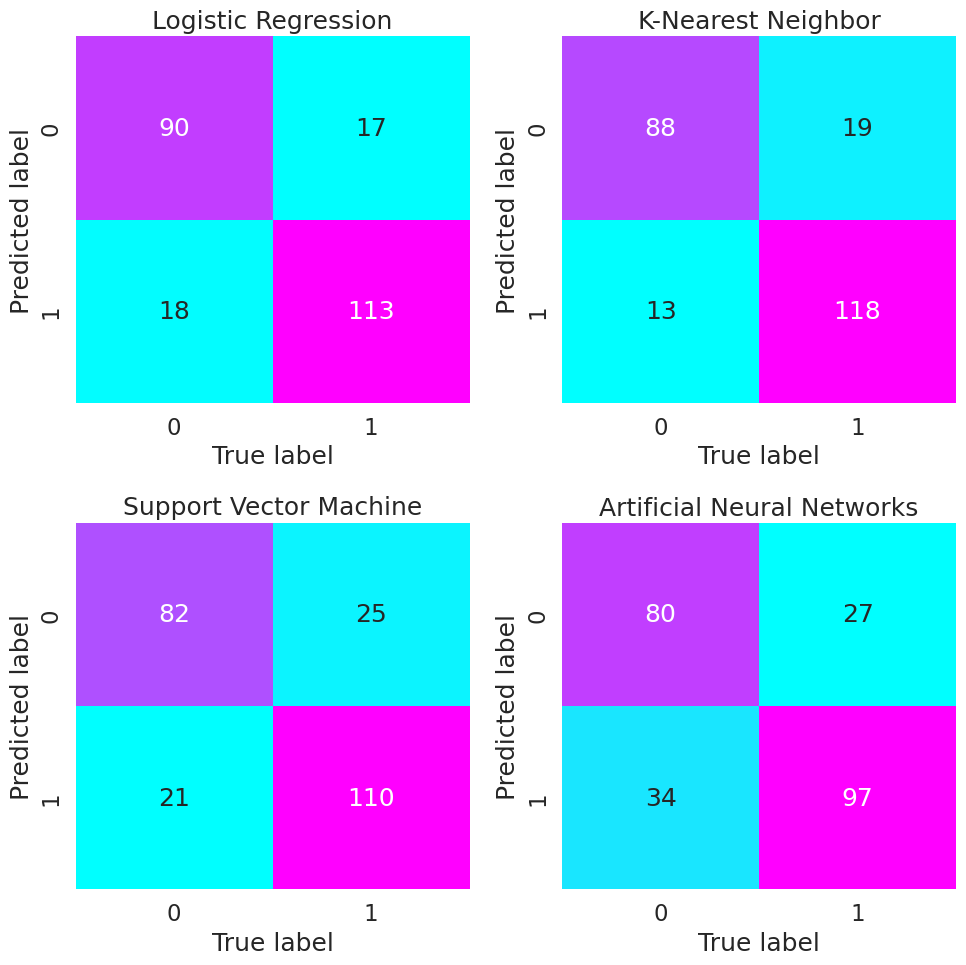}
    \end{figure}

\noindent
We see from the above comparison that KNN seems to perform better in terms of true positive identification, but Logistic Regression gives a more balance performance across both classes. ANN may need further optimization to have similar performances.
\newpage
\section{Conclusion}
\noindent
There are many important discoveries on these heart disease’s predictors. Age, resting blood pressure and maximum heart rate have moderate correlations with the presence of heart diseases which indicate their importance as predictors. On the other hand, cholesterol has weak correlations with other variables suggesting that it may have a little predictive power in the model. It is important to note that high cholesterol is a significant risk factor for heart disease despite the weak correlations between the dataset analysis and cholesterol. Mayo Clinic mentions that elevated low-density lipoprotein (LDL) cholestrol gradually causes plaques in the arteries which can cause chest pain, heart attacks and strokes \citep{mayoclinic2023cholesterol}. Therefore, it is crucial to take into account external clinical knowledge and deal with potential data problems so as to properly understand the actual relationship between cholesterol and heart disease risk.\\

\noindent
Performance evaluations of different models reveal that Logistic Regression and K-Nearest Neighbor (KNN) had shown strong overall performances where KNN was more effective in identifying positive class especially. Support Vector Machine (SVM) also showed balanced performance but higher false positives and false negatives than Logistic Regression and KNN. Artificial Neural Networks (ANN) were the least effective performers among all and might need some fine tuning or additional data for better accuracy. Moreover, there are more cases of heart disease in this dataset making it imbalanced; affecting the model’s results negatively\\

\section{Recommendations}
\noindent
Further study may want to explore log or Box-Cox for normalisation of variables like resting blood pressure, cholestrol and oldpeak in order to tackle their skewness. Techniques like SMOTE (Synthetic Minority Over-sampling Technique) can be employed to handle the class imbalances. \\

\noindent
In an effort to improve model performance , the data used can be increased with further exploration of hyper parameters to be tuned. Ensemble methods such as Random Forests or Gradient Boosting can be included for better performance\\

\noindent
Conducting feature selection techniques like Recursive Feature Elimination (RFE) will help identify and retain features with the most impact.\\

\noindent
Metrics like the ROC-AUC score should be employed to assess model performance beyond the performance metrics used in this study

\subsection*{Data Availability}
\noindent
The data used to support the findings of this study is available on Kaggle. \href{https://www.kaggle.com/datasets/sid321axn/heart-statlog-cleveland-hungary-final}{Heart Disease Dataset on Kaggle}
\subsection*{Conflict of Interest}
\noindent
The authors declare that there are no conflicts of interest.
\subsection*{Acknowledgment}
\noindent
The authors thank University of Ghana Actuarial Science and Statistics Department,Department of Computer Science and the School of Mathematical and Statistical Sciences at University of Texas Rio Grande Valley.

\end{document}